\documentclass[letterpaper]{article} % DO NOT CHANGE THIS
\usepackage{aaai2026}  % DO NOT CHANGE THIS
\usepackage{times}  % DO NOT CHANGE THIS
\usepackage{helvet}  % DO NOT CHANGE THIS
\usepackage{courier}  % DO NOT CHANGE THIS
\usepackage[hyphens]{url}  % DO NOT CHANGE THIS
\usepackage{graphicx} % DO NOT CHANGE THIS
\usepackage{natbib}  % DO NOT CHANGE THIS AND DO NOT ADD ANY OPTIONS TO IT
\usepackage{caption} % DO NOT CHANGE THIS AND DO NOT ADD ANY OPTIONS TO IT
\usepackage{algorithm}
\usepackage{algorithmicx}
\usepackage{amsmath}
\usepackage{amssymb}
\usepackage{amsfonts}
\usepackage{newfloat}
\usepackage{listings}
\DeclareCaptionStyle{ruled}{labelfont=normalfont,labelsep=colon,strut=off} % DO NOT CHANGE THIS
\floatstyle{ruled}
\newfloat{listing}{tb}{lst}{}
\floatname{listing}{Listing}
\usepackage{array}
\usepackage{booktabs}
\usepackage{algpseudocode}

\title{Harnessing Rule-Based Reinforcement Learning for Enhanced Grammatical Error Correction}
\nocopyright
\author{
    Yilin Li\textsuperscript{\rm 1},
    Xunjian Yin\textsuperscript{\rm 1},
    Yilin Chen\textsuperscript{\rm 2},
    Xiaojun Wan\textsuperscript{\rm 1}\thanks{Corresponding author.}
}
\affiliations{
    \textsuperscript{\rm 1}Wangxuan Institute of Computer Technology, Peking University\\
    \textsuperscript{\rm 2}School of Computer Science \& Key Lab of High Confidence Software Technologies (MOE), Peking University\\
    liyilin25@stu.pku.edu.cn
}

\usepackage{bibentry}
\begin{document}

\maketitle

\begin{abstract}
Grammatical error correction is a significant task in NLP. Traditional methods based on encoder-decoder models have achieved certain success, but the application of LLMs in this field is still underexplored. Current research predominantly relies on supervised fine-tuning to train LLMs to directly generate the corrected sentence, which limits the model's powerful reasoning ability. To address this limitation, we propose a novel framework based on  Rule-Based RL. Through experiments on the Chinese datasets, our Rule-Based RL framework achieves \textbf{state-of-the-art }performance, with a notable increase in \textbf{recall}. This result clearly highlights the advantages of using RL to steer LLMs, offering a more controllable and reliable paradigm for future development in GEC.
\end{abstract}


\section{Introduction}

Grammatical Error Correction (GEC)~\cite{bryant-etal-2023-grammatical} is a fundamental task in Natural Language Processing (NLP) focused on the automatic detection and correction of grammatical errors in text. This task is essential not only for improving text quality but also for supporting applications such as language learning and automated writing evaluation. Over the years, numerous models have been developed to address GEC, including the Transformer model~\cite{junczys-dowmunt-etal-2018-approaching}, BERT~\cite{kaneko-etal-2020-encoder}, and T5~\cite{rothe-etal-2021-simple}. \citet{qorib-etal-2022-frustratingly-easy-system-combination} combines these models and generates better corrections. 

\begin{figure}[htbp!]
    \centering
    \includegraphics[width=1\linewidth]{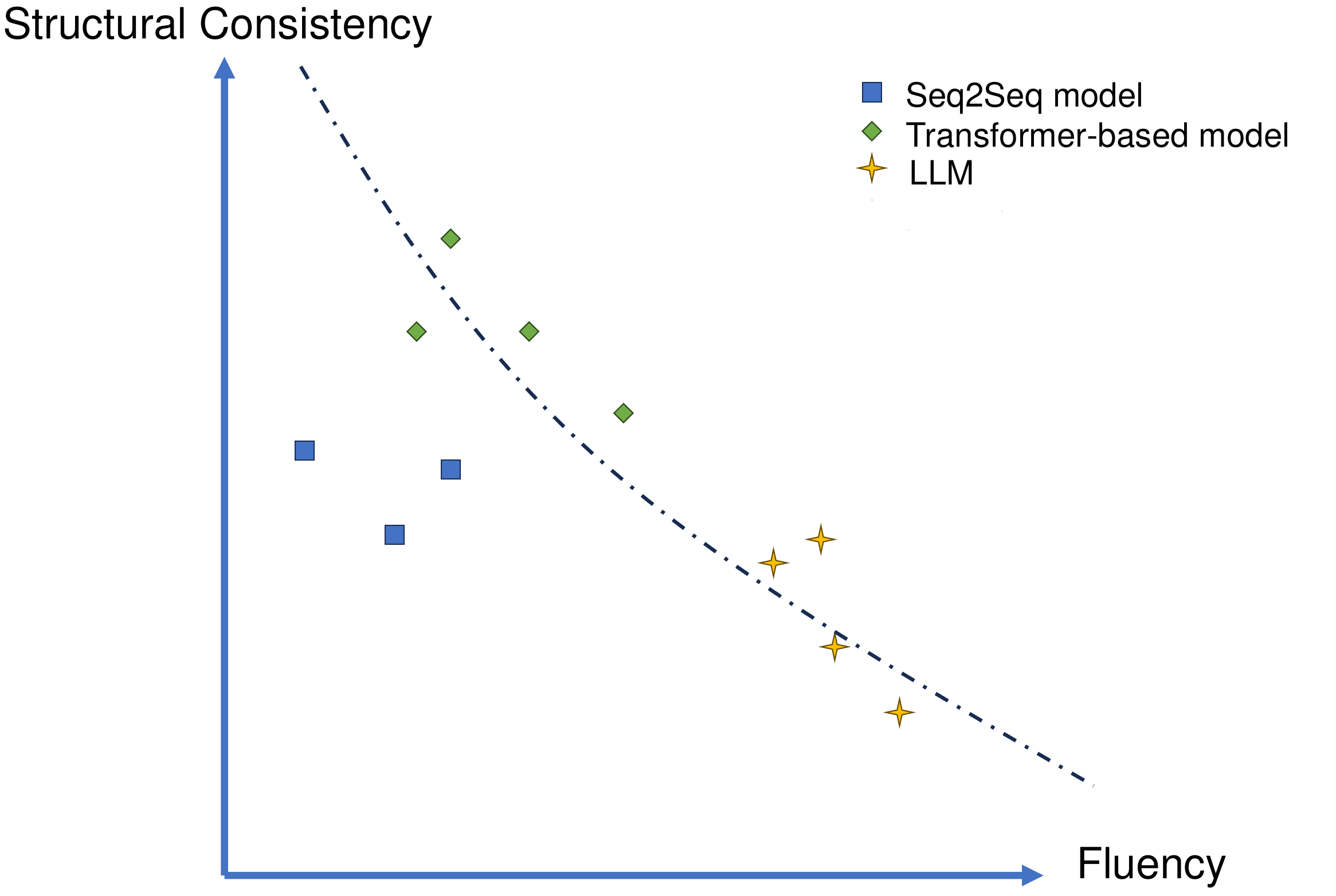}
    \caption{Traditional Seq2Seq and transformer-based models with supervised learning in GEC task prioritize precision, making fewer corrections to sentence structure. In contrast, LLMs emphasize grammar and fluency, leading to deeper corrections but often causing over-correction.}
    \label{fig:first}
\end{figure}
The advent of Large Language Models (LLMs) has markedly advanced the field of NLP. Models such as GPT and LLaMA\cite{openai2024gpt4technicalreport,grattafiori2024llama3herdmodels} have demonstrated remarkable performance and potential in various NLP tasks, attributable to their ability to discern complex syntactic, semantic, and contextual patterns. Considerable research has investigated the potential of LLMs in GEC. \citet{fang2023chatgpthighlyfluentgrammatical} and \citet{loem-etal-2023-exploring} have examined the performance of large language models in the GEC task, demonstrating that LLMs possess strong capabilities in capturing syntactic and semantic nuances.

\begin{table*}[ht!]
\centering
\begin{tabular}{@{}>{\raggedright}p{3cm}>{\raggedright\arraybackslash}p{10cm}@{}}
\toprule
\textbf{Description} & \textbf{Sentence} \\
\midrule
Source Sentence & My advice to any one start learn this sport to become carefully...  \\
\midrule
One Possible Standard Answer. & My advice to \textbf{anyone} \textbf{starting} \textbf{learning} this sport \textbf{is} to become \textbf{careful}... \\
\midrule
LLM & My advice to \textbf{anyone} \underline{\textbf{who is}} \textbf{starting to learn} this sport \textbf{is} to \underline{\textbf{be}} \textbf{careful}... \\

\bottomrule
\end{tabular}
\caption{An example demonstrating the overcorrection by large language models shows that when faced with a sentence with grammatical error, LLMs make unnecessary adjustments to the original sentence for issues like fluency. This may even bring the risk of changing the meaning of the sentence.}
\label{tab:sentence_correction}
\end{table*}
However, initial studies suggest that these LLMs struggle to surpass traditional Seq2Seq models on the GEC task\cite{qu2025evaluatingcapabilitylargescalelanguage,zhang2023multitaskinstructiontuningllama,yang-quan-2024-alirector}. One prominent issue is overcorrection, where grammatically correct text segments are unnecessarily modified, thereby compromising the integrity of the original sentence. As shown in Figure \ref{fig:first}, traditional methods with supervised learning can carefully ensure consistency in the form of input and output text but often lead to missed error corrections, whereas large models tend to ambitiously overcorrect to make sentences fluent. Simple prompting techniques fail to ensure that LLMs remain faithful to the original text, leading to a trade-off between fluency and structural fidelity\cite{sun-wang-2022-adjusting}.

Table~\ref{tab:sentence_correction} provides an example of overcorrection, highlighting how an LLM introduces unnecessary modifications to a sentence.

The prevailing strategy among researchers is to treat LLMs as generative models, employing Supervised Fine-Tuning (SFT) to enable them to produce corrected sentences directly. However, this approach may not fully harness the inherent reasoning capabilities of LLMs. Efforts to integrate chain-of-thought (CoT) reasoning into LLMs for GEC have been explored, yet these often lead to issues such as hallucinations, deviations from task instructions, and a heightened risk of overcorrection. According to~\citet{fang2023chatgpthighlyfluentgrammatical}, CoT has proven less effective, possibly due to the insufficient reasoning capacity of LLMs to address the intricacies of GEC. Simply embedding reasoning techniques like CoT does not mitigate these shortcomings, as even the most advanced models—both open-source and proprietary—struggle to identify errors in challenging sentences.

Recent developments, such as the introduction of OpenAI's O1\cite{openai2024openaio1card} and DeepSeek's R1\cite{guo2025deepseek}, have bolstered the reasoning abilities of LLMs. Simultaneously, research interest in long reasoning and rule-based reward mechanisms within reinforcement learning has surged. This paper investigates the integration of rule-based rewards into GEC. We also assess the performance of the DeepSeek R1 and other reasoning models on GEC tasks. Using DeepSeek R1, we generated training data enriched with extended reasoning processes. Subsequently, we fine-tuned the model with supervised learning on this data, followed by reinforcement learning incorporating rule-based rewards to enhance its GEC performance.

The contributions of this paper are as follows:
\begin{itemize}
    \item We leverage LLMs as reasoners for GEC, marking the first exploration of models with enhanced reasoning capabilities in this context.
    \item We apply reinforcement learning to GEC by designing a rule-based reward function specifically designed for Grammatical Error Correction.
    \item We develop a Chinese GEC model using this methodology, which surpasses existing baselines on the FCGEC dataset, delivering more accurate and interpretable corrections. Additionally, we have made the code and train dataset generated by DeepSeek R1 publicly available.
\end{itemize}

\section{Related Work}
\subsection{Traditional GEC Methods}

Traditional GEC methods can be divided into two categories: Sequence-to-Edit(Seq2Edit) and Sequence-to-Sequence (Seq2Seq). 

\noindent\textbf{Seq2Seq}\quad Early work primarily focuses on sequence-to-sequence models \cite{junczys-dowmunt-etal-2018-approaching}, which treats GEC as a translation task, translating erroneous sentences into corrected ones. Enhancements such as data synthesis and advanced reranking strategies have further improved these models \cite{stahlberg-kumar-2021-synthetic, lichtarge2020dataweightedtrainingstrategies} More advanced Seq2Seq approaches use Transformer-based models. Transformer-based models have played a crucial role in recent developments, leveraging architectures like BERT, BART and T5 \cite{ tarnavskyi-etal-2022-ensembling-sequence-taggers,2019bart, Raffel2019ExploringTL}, which excel at handling long dependencies. These models have been fine-tuned on GEC-specific datasets, achieving state-of-the-art results. Pre-training strategies and large-scale unsupervised data have been instrumental in this improvement \cite{grundkiewicz2019neural}.

\noindent\textbf{Seq2Edit}\quad The Seq2Edit approach frames GEC as a sequence labeling task by predicting the appropriate edit operation for each token. Models like GECToR \cite{omelianchuk-etal-2020-gector1}, have since gained prominence, introducing an efficient token-level correction process that tags errors instead of rewriting entire sentences. This model reduces inference time while maintaining high accuracy, particularly in low-resource settings \cite{stahlberg-kumar-2020-seq2edits}.

\subsection{LLMs for GEC}

LLMs such as GPT-3 and GPT-4 have been employed for GEC \cite{fang2023chatgpthighlyfluentgrammatical}, although they face challenges related to over-correction. Recent studies indicate that these models perform well when guided with in-context examples \cite{tang2024ungrammaticalsyntaxbasedincontextexampleselection}. \citet{tang2024ungrammaticalsyntaxbasedincontextexampleselection} uses syntactic information to select in-context examples. 

In another line, some research have explored
other roles of LLMs in the GEC task, such as generating explanations for corrections, data augmentation\cite{li2024rethinkingroleslargelanguage,song-etal-2024-gee,wang2024improvinggrammaticalerrorcorrection} and assessing the quality of grammatical edits\cite{Xie_2025}.

% \begin{figure*}[htbp!]
%     \centering
%     \includegraphics[width=1\linewidth]{main.png}
%     \caption{Enter Caption}
%     \label{fig:enter-label}
% \end{figure*}

\subsection{LLM Reasoning with Post-training.}
Previous work has primarily relied on supervised fine-tuning with carefully curated datasets to enhance LLM performance in complex tasks like reasoning or tool use~\cite{schick2023toolformer, qin2023toolllm, gou2024toratoolintegratedreasoningagent}. 
Recently, reinforcement learning has gained traction as a more scalable and generalizable training paradigm. 
The development of RL methods for LLMs has evolved from reinforcement learning from human feedback (RLHF)~\cite{kaufmann2023survey} and proximal policy optimization (PPO)~\cite{schulman2017proximal} to more advanced techniques such as direct preference optimization (DPO)~\cite{rafailov2023direct}, SimPO~\cite{meng2024simpo}, and group relative policy optimization (GRPO)~\cite{shao2024deepseekmath}. 

Among these, GRPO~\cite{shao2024deepseekmath} is specifically designed for LLMs, replacing the traditional critic with a group-based evaluation strategy. 
It has shown strong performance in enhancing reasoning abilities across a range of tasks, including mathematical problem solving~\cite{shao2024deepseekmath, xie2025logic}, search engine interaction~\cite{jin2025search, song2025r1}, and code generation~\cite{li2025torl}. 
A pivotal application of this algorithm was demonstrated by the open-source community with DeepSeek-R1~\cite{guo2025deepseek}, which showed that large-scale pure RL guided only by simple rule-based rewards (i.e., formatting rules and final answer correctness) can motivate LLMs to develop self-emergent reasoning processes. 
This "R1-Zero" paradigm has been successfully replicated and extended to other domains, including logic games~\cite{xie2025logic} and vision reasoning~\cite{huang2025vision}. 
The flexibility of GRPO's reward function has also been leveraged for diverse objectives, such as assigning weights to sub-tasks~\cite{yu2024steptool} or constraining tool use frequency~\cite{li2025torl}.

In this work, we build upon the established success of the GRPO algorithm and extend its application to GEC.

\section{Similarities between Grammatical Error Correction and Math Reasoning Tasks}

Grammatical Error Correction can be viewed as a complex reasoning task, sharing significant parallels with mathematical reasoning due to its reliance on multi-level rule comprehension and structured thought processes.

First, both GEC and mathematical reasoning demand a deep understanding of underlying rules. Just as math relies on precise laws and logic, GEC requires applying grammatical principles (e.g., subject-verb agreement, tense consistency, word usage) to identify and correct errors. This shared need for rule adherence is a key similarity.

Second, both tasks require a structured, step-by-step reasoning process. Mathematics often involves decomposing complex problems and solving them incrementally. Similarly, GEC involves analyzing sentences, identifying errors, and deducing corrections systematically, especially for multiple errors within a sentence, much like stepwise mathematical problem-solving.

Third, both GEC and math reasoning operate with clear objectives and evaluation criteria. Math aims for accurate, logically sound solutions. GEC seeks to produce grammatically correct and natural sentences, so we can compare outputs with the correct sentences by means of string comparison, and then give reward scores.

Therefore, viewing GEC as a complex reasoning task clarifies its underlying logic and suggests new paths for improvement. This perspective allows for adapting systematic problem-solving approaches from mathematics to enhance the accuracy and efficiency of GEC.

\section{Our Methodology}

\begin{algorithm*}[htbp!]
\caption{Two-Stage Data Generation and Filtering for SFT}
\label{alg:data_generation_compact}
\begin{algorithmic}[1]
\State \textbf{Input:} $D_{Source1}, D_{Source2}$ (Source datasets for Stages 1 and 2)
\State \textbf{Input:} $M_{Qwen}$ (Qwen3-32B), $M_{R1}$ (DeepSeek-R1), $M_{V3}$ (DeepSeek-V3)
\State \textbf{Output:} $D_{SFT1}, D_{SFT2}$ (High-quality datasets for SFT Stages 1 and 2)
\Statex
\State \Comment{--- \textbf{Stage 1: Initial Data Generation} ---}
\State $D_{SFT1} \gets \emptyset$
\For{each sentence $s$ in $D_{Source1}$}
    \State $s_{reasoning} \gets M_{Qwen}(s)$
    \State Add $(s, s_{reasoning})$ to $D_{SFT1}$
\EndFor
\Statex
\State \Comment{--- \textbf{Stage 2: Iterative Generation and Filtering} ---}
\State $D_{SFT2} \gets \emptyset$
\For{each sentence $s$ in $D_{Source2}$}
    \State $s_{reasoning_1} \gets M_{R1}(s)$ \Comment{First generation pass}
    \If{$M_{V3}$ accepts $s_{reasoning_1}$}
        \State Add $(s, s_{reasoning_1})$ to $D_{SFT2}$
    \Else
        \State $s_{reasoning_2} \gets M_{R1}(s)$ \Comment{Second generation pass for failed instances}
        \If{$M_{V3}$ accepts $s_{reasoning_2}$}
            \State Add $(s, s_{reasoning_2})$ to $D_{SFT2}$
        \EndIf
    \EndIf
\EndFor
\Statex
\State \Return $D_{SFT1}, D_{SFT2}$
\end{algorithmic}
\end{algorithm*}

This section details our methodology, which trains a GEC model through a two-stage process: an initial SFT phase followed by a reinforcement learning RL phase. We design a rule-based reward function specifically for the GEC task. This function integrates two key signals: a reward for adhering to the correct reasoning format and a reward for the final answer's correctness. We use this composite reward to train the model with the GRPO algorithm~\cite{shao2024deepseekmath}, which ensures stable and efficient RL training.

\subsection{Data Generation}
\begin{table}[htbp!]
\centering

\setlength{\tabcolsep}{4pt} 

\begin{tabular}{lrrp{2.5cm}} % Use 'p' column for automatic text wrapping
\hline
\textbf{Dataset} & \textbf{\#Sents} & \textbf{\%Error} & \textbf{Usage} \\
\hline\hline
\multicolumn{4}{l}{\textit{Training Sets}} \\
Lang8 & 1,220,906 & 89.5 & SFT Stage I \\
HSK & 156,870 & 60.8 & SFT Stage I \\
FCGEC & 36,341 & 54.3 & SFT Stage II \& RL Training \\
\hline
\multicolumn{4}{l}{\textit{Validation Set}} \\
FCGEC-dev & 2,000 & 55.1 & Validation \\
\hline
\multicolumn{4}{l}{\textit{Test Sets}} \\
FCGEC-test & 3,000 & -- & Testing \\
NaCGEC-test & 5,869 & 95.6 & Testing \\
\hline

\end{tabular}
\caption{Statistics of the used datasets. \#Sentences denotes the number of the sentences and \% Error denotes the percentage of the erroneous sentences.}
\label{tab:chinese_datasets_single_col}
\end{table}

Following previous work \cite{zhang-etal-2022-syngec}, our Supervised Fine-Tuning process is divided into two stages, with the corresponding datasets detailed in Table \ref{tab:chinese_datasets_single_col}. For SFT-stage 1, we adopt the data preparation methodology from \citet{zhang-etal-2022-syngec}. Specifically, all error-free samples are discarded from the Lang8 and HSK datasets. The HSK dataset is then replicated five times and combined with the Lang8 dataset, yielding a total of 1,568,885 sentence pairs. To create the data for SFT-stage 2, we use the FCGEC training set. The full procedure is detailed in Algorithm \ref{alg:data_generation_compact}.

We leverage Qwen-32B and DeepSeek-R1 to perform inference on the SFT-stage 1 and SFT-stage 2 datasets, respectively, to generate a new corpus containing detailed reasoning traces. Subsequently, we employ DeepSeek-V3 to perform quality filtering on the data generated for SFT-stage 2. The objective of this filtering is to select instances where the reasoning path correctly identifies whether a sentence needs correction and whether the proposed edit is accurate. The prompts used for generate detailed reasoning traces and the filtering process are detailed in the Appendix. Instances that fail this initial check are then re-generated using R1 in a second generation pass. This refined data is filtered again. After this filtering and refinement process, we ultimately obtained 27,501 high-quality, reasoning-augmented GEC training instances. Meanwhile, the original FCGEC training set is retained for the RL stage.

\subsection{Rule-Based Reward}

In RL, the reward is the main signal that drives model training. DeepSeek-R1-Zero \citep{guo2025deepseek} employs simple rule-based rewards that check whether the final answer is correct and whether the response follows a specific format. This works well for tasks with fixed format correct answers such as math or coding. For GEC, we can also use such a simple method for rewards. Since there might be multiple ways to correct a grammatical error, we can grant the full reward for the entire answer as long as one of the valid corrections is met.

\begin{figure}
    \centering
    \includegraphics[width=0.9\linewidth]{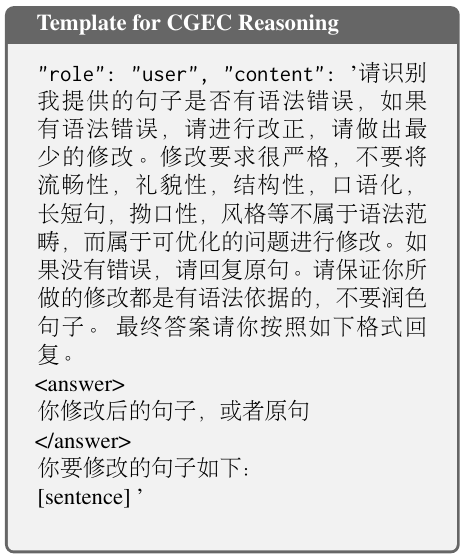}
    \caption{Template for CGEC Reasoning}
    \label{fig:pmt}
\end{figure}

We use a structured prompt template similar to that in DeepSeek-R1-Zero in Fig\ref{fig:pmt}. We use the same \texttt{<think>} tag format as Qwen3 for data generation, so our prompt do not need to specify this formatting requirement. The English translation of this prompt can be found in the Appendix.

Our comprehensive reward function, $R_{\text{total}}$, is designed to optimize model outputs for both structural integrity and semantic accuracy. It is a sum of two components: a Rule Reward ($R_{\text{rule}}$) and a Correctness Reward ($R_{\text{c}}$).

\paragraph{Rule Reward ($R_{\text{rule}}$)}
$R_{\text{rule}}$ combines rewards for correct usage of predefined structural tags: open tag $S_o$ (\texttt{<answer>}), close tag $S_c$ (\texttt{</answer>}) with a penalty for excess content length ($L_{\text{suffix}}$) appearing after a specific delimiter $S_d$(\texttt{</answer>}).
\begin{equation}
\label{eq:r_rule}
\begin{split}  
R_{\text{rule}}(T) = {} & +0.125 \cdot \mathbb{I}(\text{count}(S_o, T)=1) \\  % 
                      & +0.125 \cdot \mathbb{I}(\text{count}(S_c, T)=1) \\   
                      & -0.001 \cdot \mathbb{I}(\text{count}(S_c, T)=1) \cdot L_{\text{suffix}}(T, S_d)
\end{split} 
\end{equation}
Here, $\mathbb{I}(\cdot)$ is the indicator function signifying presence of the respective tags. $S_o$, $S_c$, and $S_d$ are specific predefined string constants. $L_{\text{suffix}}(T, S_d)$ measures the length of content trailing the delimiter $S_d$; this penalty component is applied only if the close tag $S_c$ is present. The coefficient for the rule-based reward is set low, as the model has already been refined through a two-stage SFT process before RL training. This reward therefore serves as a minimal penalty aimed only at preventing formatting errors or truncation from repetitive outputs.

\paragraph{Correctness Reward ($R_{\text{c}}$)}
$R_{\text{c}}$ quantifies the semantic accuracy of the model's extracted answer $R$, evaluated against the original input sentence $Q$ and the set of ground truth answers $A$. The design of our correctness reward function is guided by the evaluation metric $F_{0.5}$ score and thus places a strong emphasis on precision. 
To reflect this, we significantly reward the model for correctly preserving an error-free sentence. 

We experimented with reward values of 4 and 6 for this case, ultimately selecting 6 as it yielded a higher $F_{0.5}$ score.
To encourage corrections of erroneous sentences, we assign a process reward of 0.1 for instances where the model modifies an incorrect sentence, even if the modification itself is still incorrect. For a correct modification of an erroneous sentence, the model receives a base reward of 2, totaling 2.1 with the process reward.

Conversely, to penalize inaction and over-correction, we designed distinct penalties: -0.05 for failing to modify an incorrect sentence, and a harsher -0.1 for incorrectly modifying a correct one. 
During our experiments, we tested a uniform penalty of -0.1 for both scenarios but found that the differentiated penalties of -0.05 and -0.1 yielded superior performance. 
Therefore, we adopted this latter configuration, which aligns with our focus on prioritizing precision.
\begin{equation}
R_{\text{c}}=
\begin{cases}
4.0, & \text{Original correct, model preserved.} \\
2.1, & \text{Original incorrect, model corrected.} \\
0.1, & \text{\shortstack{Original incorrect, model changed,\\ still incorrect.}} \\
-0.05, & \text{Original incorrect, model unchanged.} \\
-0.1, & \text{Original correct, model changed.}
\end{cases}
\end{equation}

\subsection{RL Algorithm}

We use the GRPO algorithm with Clip-Higher strategy~\citep{shao2024deepseekmath} to train the  LLMs with our rule reward. In each training step, for a given sentence  $q$, we sample a group of candidate outputs $\{o_1,o_2,\cdots,o_G\}$ from the policy model $\pi_{\theta_{old}}$. $A_i = \frac{r_i - \operatorname{mean}(\{r_1, r_2, \dots, r_G\})}{\operatorname{std}(\{r_1, r_2, \dots, r_G\})}$ is the computed advantage using the group rule-metric mixed rewards $\{r_1,r_2,\cdots,r_G\}$. GRPO then maximizes the following objective function to optimize $\pi_{\theta}$:  
\begin{equation}
\begin{aligned}
J_{\mathrm{GRPO}}(\theta) 
&= \mathbb{E}_{q \sim P(Q),\, \{o_i\}_{i=1}^G \sim \pi_{\theta_{\mathrm{old}}}(O \mid q)} \\
&\Biggl[
  \frac{1}{G} \sum_{i=1}^G
  \min\!\Bigl(
    \frac{\pi_{\theta}(o_i \mid q)}{\pi_{\theta_{\mathrm{old}}}(o_i \mid q)}\,A_i,\, \\
    &\mathrm{clip}\!\Bigl(
      \frac{\pi_{\theta}(o_i \mid q)}{\pi_{\theta_{\mathrm{old}}}(o_i \mid q)},
      1-\varepsilon_{low},\,
      1+\varepsilon_{high}
    \Bigr)
    A_i
  \Bigr)  \\
  &-\,\beta\,D_{\mathrm{KL}}\bigl(\pi_{\theta}\,\big\|\,\pi_{\mathrm{ref}}\bigr)
\Biggr],
\end{aligned}
\label{eq1}
\end{equation}
where $\varepsilon_{low}$, $\varepsilon_{high}$ and $\beta$ are hyperparameters controlling the PPO clipping threshold and the weight of the Kullback–Leibler (KL) divergence penalty~\cite{schulman2017proximal,shao2024deepseekmath}, respectively. Specifically, $\varepsilon$ determines the permissible range for policy updates, while $\beta$ regulates the magnitude of the KL penalty during training to prevent excessive policy shifts from the reference policy $\pi_{ref}$ (typically the initialization of $\pi_{\theta}$). $D_{KL}\bigl(\pi_\theta \,\|\, \pi_{\text{ref}}\bigr)
= \frac{\pi_{\text{ref}}(o_i \mid q)}{\pi_\theta(o_i \mid q)}
- \log\!\Bigl(\frac{\pi_{\text{ref}}(o_i \mid q)}{\pi_\theta(o_i \mid q)}\Bigr)
- 1 \,$ is the KL divergence approximation term.

\begin{table*}[htbp!]
\centering
\small

\begin{tabular}{lcccccc}
\toprule
& \multicolumn{3}{c}{\textbf{FCGEC}} & \multicolumn{3}{c}{\textbf{NaCGEC}} \\
\cmidrule(lr){2-4} \cmidrule(lr){5-7}
\textbf{System} & \textbf{P} & \textbf{R} & $\mathbf{F_{0.5}}$ & \textbf{P} & \textbf{R} & $\mathbf{F_{0.5}}$ \\
\midrule
\multicolumn{7}{l}{\textit{Traditional GEC Baselines}} \\
GECToR~\cite{omelianchuk-etal-2020-gector1} & 46.11 & 34.35 & 43.16 & - & - & - \\
BART~\cite{2019bart} & 38.38 & 37.62 & 38.23 & 62.04&45.84&57.94  \\
SynGEC~\cite{zhang-etal-2022-syngec} & 63.75 & \underline{39.78} & 56.89 & \underline{62.42}&\underline{47.41}&\underline{58.71} \\
LM-Combiner~\cite{wang2024lmcombinercontextualrewritingmodel} & 55.67 &39.04 &51.30 & - & - & - \\
MrGEC~\cite{liu-etal-2024-towards-better} & \underline{\textbf{65.71}} &37.78 &\underline{57.22} & - & - & - \\
\midrule
\multicolumn{7}{l}{\textit{LLM-based Baselines}} \\
Instruction Tuning~\cite{liu-etal-2025-chain} & \underline{65.65} &36.49 &\underline{56.60}& 62.50 &40.72 &56.46 \\
Alirector~\cite{yang-quan-2024-alirector} & 64.49 & 36.22 & 55.78 & \underline{\textbf{66.04}} &\underline{45.91}& \underline{\textbf{60.71}} \\
De-CoGLM~\cite{li2024detectioncorrectionstructuregenerallanguage} & 56.09 &\underline{38.02} &51.22 & - & - & - \\
\midrule
\multicolumn{7}{l}{\textit{Reasoning Models}} \\
O3 Mini~\cite{qu2025evaluatingcapabilitylargescalelanguage} & 8.24 &19.44 &9.31 &8.59 &21.66 &9.77 \\
QWQ 32B~\cite{qu2025evaluatingcapabilitylargescalelanguage} & 17.33 &32.15 &19.09& 17.55& 33.21& 19.38\\
R1~\cite{qu2025evaluatingcapabilitylargescalelanguage} & \underline{18.78} &\underline{36.06} &\underline{20.77} &\underline{19.31} &\underline{37.30}& \underline{21.37} \\
% Doubao-1.5-pro~\cite{qu2025evaluatingcapabilitylargescalelanguage} & \underline{32.18} &\underline{45.08} &\underline{34.14} &\underline{35.93} &\underline{\textbf{51.53}} &\underline{38.25} \\
\midrule
\multicolumn{7}{l}{\textit{Ours}} \\
SFT (Direct Generation) & \underline{64.81}	&36.79	&56.24 & \underline{64.01}  &43.01  &58.32 \\
SFT (with Reasoning) & 57.96&	46.15&	55.14 & 47.91 & 41.72 & 46.53 \\
~~+ 16 vote & 58.02&	40.97&	53.56 & 48.40  & 40.94 & 46.70 \\
RL (with Reasoning)& 60.68	&46.95	&57.33 & 61.97 & 47.88 &58.52 \\
~~+ 16 vote & 61.84 &\underline{\textbf{48.94}}	&\underline{\textbf{58.74}} & 61.94 &\underline{\textbf{49.68}} &\underline{59.03} \\
\bottomrule
\end{tabular}

\caption{Comparison of our method with baselines on the FCGEC and NaCGEC test sets. The best result in each category is \underline{underlined}, and the overall best result among all baselines is in \textbf{bold}.}
\label{tab:main_results_combined}
\end{table*}

\section{Experiments}

\subsection{Experimental Setup}
\label{sec:exp_set}

\noindent{\textbf{Dataset and Benchmarks.}}
Following the setup of \citet{liu-etal-2025-chain}, we use the Chinese native-speaker datasets FCGEC\cite{xu-etal-2022-fcgec} (in-domain) and NaCGEC\cite{ma2022linguistic} (out-of-domain), which are challenging datasets featuring errors and linguistic complexities typical of native speakers. This strategic focus ensures the authentic correction of in-distribution erroneous sentences, mitigating common OOD problems from L2 learner texts, which exhibit distinct error patterns. We focus on Chinese to avoid the evaluation challenges posed by English L2 benchmarks like BEA-19\cite{bryant-etal-2019-bea-19} and CoNLL-14\cite{ng-etal-2014-conll14}. On these datasets, the strong tendency of LLMs to rewrite sentences for fluency conflicts with evaluation metrics that penalize unnecessary edits, often resulting in low scores.

\noindent{\textbf{Evaluation Metrics.}}
For the validation set experiments and for the test set NaCGEC, we use the official evaluation tool ChERRANT
\footnote{\url{https://github.com/HillZhang1999/MuCGEC/tree/main/scorers/ChERRANT}}
to evaluate the model based on correction span's P/R/F0.5.
As for the test set FCGEC, we obtain the same evaluation metrics by submitting the system results in 
CodaLab
\footnote{\url{https://codalab.lisn.upsaclay.fr/competitions/8020}}
online platform.

\noindent{\textbf{Baselines}}
We compare our approach with the following groups of baselines to ensure a comprehensive evaluation.

\begin{itemize}
    \item \textbf{Traditional GEC Baselines:} We select several established GEC models for comparison. These include GECToR~\cite{omelianchuk-etal-2020-gector1}, as a representative of Seq2Edit methods, as well as the Seq2Seq methods BART~\cite{2019bart}, SynGEC~\cite{zhang-etal-2022-syngec}, LM-Combiner~\cite{wang2024lmcombinercontextualrewritingmodel}, and MrGEC~\cite{liu-etal-2024-towards-better}.

    \item \textbf{LLM-based Baselines:} To benchmark against Large Language Model approaches, we first establish two baselines trained via Supervised Fine-Tuning using our own configuration. The first is trained to directly generate the corrected sentence, while the second is trained to output the sentence after a reasoning process. Additionally, we compare our results against other prominent LLM-based methods, including Instruction Tuning\cite{liu-etal-2025-chain}, Alirector\cite{yang-quan-2024-alirector}, and DeCoGLM\cite{li2024detectioncorrectionstructuregenerallanguage}.

    \item \textbf{Reasoning Models:} Finally, to evaluate our model against those with enhanced reasoning capabilities, we cite the performance of O3 Mini, QWQ 32B, and R1 as reported in~\cite{qu2025evaluatingcapabilitylargescalelanguage}.
\end{itemize}

\noindent{\textbf{Training Details.}} 
Our Supervised Fine-Tuning process is conducted using the LlamaFactory\footnote{\url{https://github.com/hiyouga/LLaMA-Factory}}. We select Qwen3-8B as the starting model. For training on SFT-stage 1, we configure a batch size of 128 and employ a learning rate of 1e-5. The model is trained for 3 epochs , which takes approximately 20 hours to complete. For training on SFT-stage 2, we configure a batch size of 64 and employ a learning rate of 1e-5. The model is trained for 2 epochs , which take approximately 2 hours to complete.

Our implementation on RL is based on the verl\footnote{\url{https://github.com/volcengine/verl}} framework. We select the SFT model as starting models for training. During training, we configure a batch size of 128 and utilize 16 rollouts per prompt within the GRPO algorithm. We employ a constant learning rate of 1e-6 and set the sampling temperature to 1.0. The maximum generation length for responses is capped at 2000 tokens.  We set the KL penalty coefficient 
$\beta$ to 0. The PPO clipping range $\epsilon_{low}$ is set to 0.2, and $\epsilon_{high}$ is set to 0.28. All models are trained for 5 epochs for about 40 hours on A100 * 8. 

\noindent{\textbf{Inference Details.}} 
We evaluate our model under two settings. The first is a single pass using greedy decoding (t=0). The second setting employs multi-sample voting, where we generate multiple candidates with a temperature of t=1 and select the most frequent output as the final answer. We conduct this voting process with 1, 4, 8, 16, and 32 samples.

\section{Main Results}

Our experimental results, presented in Table~\ref{tab:main_results_combined}, are analyzed across both in-domain and out-of-domain test sets to provide a comprehensive evaluation of our approach.

\begin{figure*}[htbp!]
    % \centering
    \includegraphics[width=1\linewidth]{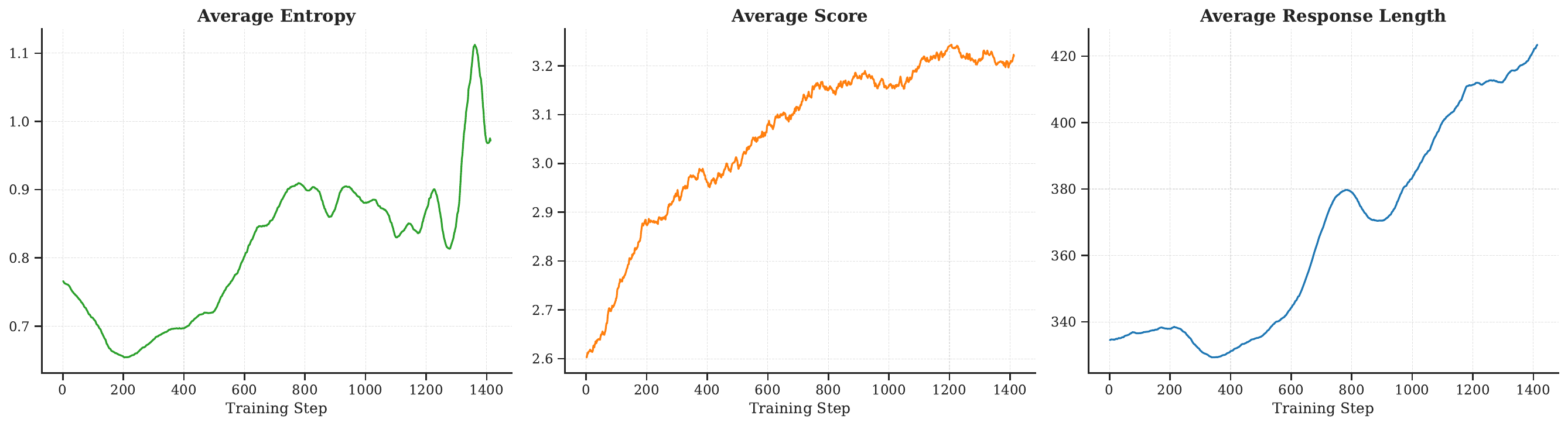}
    \caption{Training dynamics of RL process}
    \label{fig:lenth}
\end{figure*}

\begin{figure}
    \centering
    \includegraphics[width=1\linewidth]{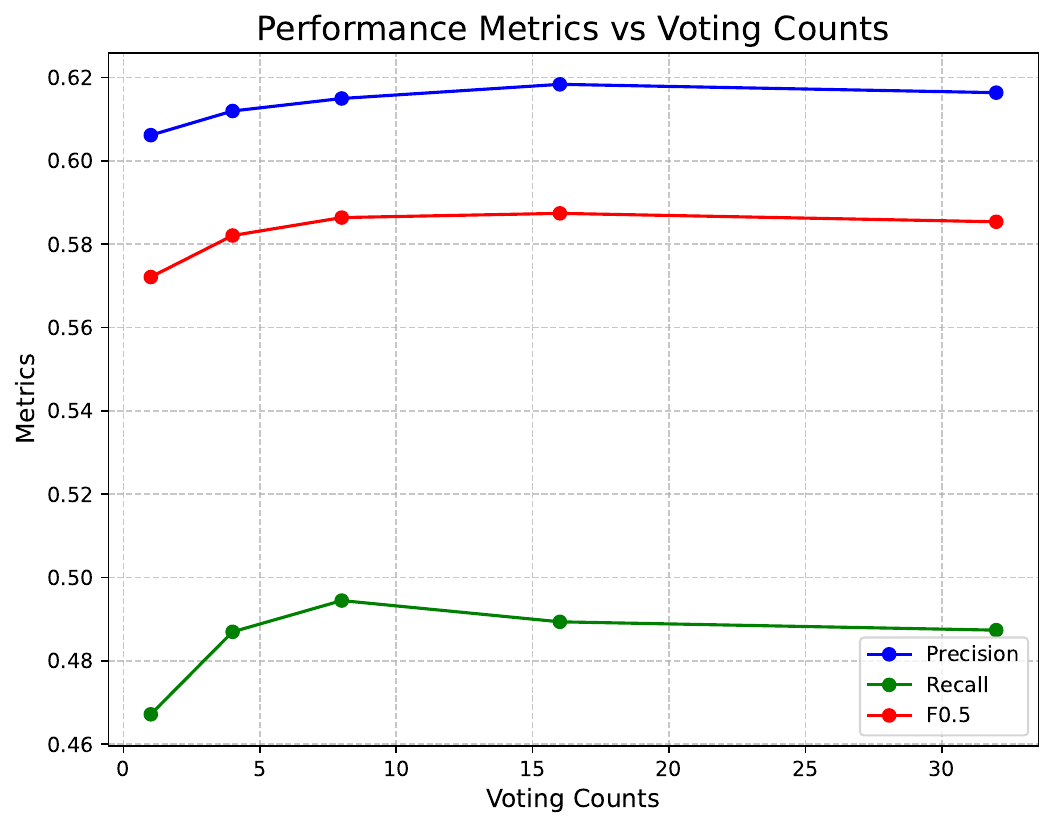}
    \caption{Vote Count and Score}
    \label{fig:vote}
\end{figure}
\subsection{In-Domain Performance (FCGEC)}

On the in-domain FCGEC test set, our findings reveal several key insights. 
First, among our self-implemented SFT baselines, a clear trade-off emerges. 
The \textbf{SFT (Direct Generation)} model achieves higher precision and a better $F_{0.5}$ score, whereas the \textbf{SFT (with Reasoning)} model demonstrates a superior recall rate. 
This suggests that while direct generation is more conservative, incorporating a reasoning process encourages the model to identify a wider range of errors, albeit at the cost of precision.

The introduction of Reinforcement Learning significantly enhances the model that utilizes reasoning. 
After RL training, the model shows marked improvements in both precision and recall. 
Ultimately, our model achieves state-of-the-art performance, surpassing all other methods in both recall and the final $F_{0.5}$ score. 
Furthermore, the effectiveness of the voting mechanism is particularly pronounced for the RL-trained model, yielding substantial gains across all metrics, which suggests that RL training improves the quality and diversity of the model's sampling process. However, the SFT model do not show improvement after voting, and even experienced a performance degradation.

When compared with other baselines, our approach demonstrates a significant advantage in recall. 
The \textbf{Reasoning Models} (O3 Mini, QWQ 32B, R1) exhibit relatively low performance, which can be attributed to their tendency to completely rewrite sentences. 
Compared to both \textbf{Traditional GEC Baselines} and other \textbf{LLM-based Baselines}, our final model's most notable improvement is its substantial increase in recall, underscoring the critical role of reasoning in enhancing error detection. This underscores the critical role of reasoning in enhancing error detection capabilities. So, to fully unlock the potential of LLMs in GEC, it is evident that we must continue to explore and effectively harness their powerful reasoning abilities.

\subsection{Out-of-Domain Generalization (NaCGEC)}

On the out-of-domain NaCGEC test set, the results highlight the robustness and generalization capabilities of our RL-based approach. 
The \textbf{SFT (Direct Generation)} model maintains a reasonable level of performance, demonstrating acceptable generalization. 
However, the \textbf{SFT (with Reasoning)} model suffers a significant performance degradation, indicating that its complex reasoning process learned via SFT is brittle and does not transfer well to out-of-domain data.

Crucially, after being trained with RL, the model's performance on the OOD dataset sees a substantial improvement. Moreover, the voting mechanism remains highly effective, further boosting performance. This combined performance lift strongly demonstrates that our RL paradigm fosters more robust and generalizable reasoning. We attribute this to the fact that the reward signal, based on the correctness of the final answer, compels the model to learn fundamental and generalizable principles of grammatical judgment, rather than superficial patterns specific to the training data distribution, thereby achieving stronger generalization.

\section{Analysis}
Figure \ref{fig:lenth} depicts the training dynamics of RL process. The second and third charts show the progression of average reward and average response length across rollout samples during the RL process, both of which steadily increase over time. The first chart presents the average entropy. Interestingly, the average entropy first decreases and then increases. We attribute this to the model initially focusing on precision to reduce severe over-correction, which lowers exploration. The latter increase in entropy indicates a shift to more exploratory behavior that boosts recall.Figure \ref{fig:vote} depicts the correlation between evaluation scores and the number of voting iterations on FCGEC test set; precision and $F_{0.5}$ scores consistently rose before stabilizing, whereas recall initially increased and then declined. Performance analysis suggests that the effectiveness of RL may stem from an enhanced quality of sampling, which allows the model to produce higher-quality responses over multiple generations.

\section{Conclusion}
In this work, we have successfully demonstrated the efficacy of applying RL with rule-based rewards to the GEC task. Our approach achieves state-of-the-art results on the in-domain FCGEC dataset, and maintains competitive performance in recall and $F_{0.5}$ score on the out-of-domain NaCGEC benchmark. This highlights the strong generalization and robustness conferred by our RL paradigm. While our method significantly enhances recall, we recognize the room for further improvement in precision. For future work, we will focus on designing a more precise reward for GEC based on the minimum edit distance.

\bibliography{aaai2026}

\newpage
\appendix
\section{Appendix}
\begin{figure}[htbp!]
    \centering
    \includegraphics[width=0.9\linewidth]{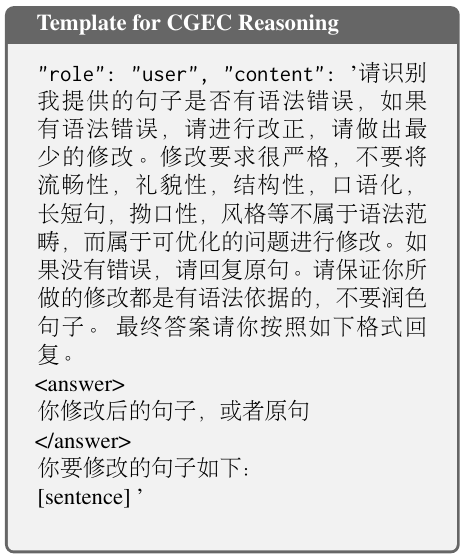}
    \caption{Template for CGEC Reasoning}
    \label{fig:pmtCGEC}
\end{figure}

The English translation of fig\ref{fig:pmtCGEC} is as follows:

Please identify if the sentence I provide has grammatical errors. If there are grammatical errors, please correct them, making the minimum necessary changes. 

The requirements for modification are very strict: do not modify for issues that fall outside the scope of grammar and are considered optimizable, such as fluency, politeness, structure, colloquialism, sentence length, awkwardness, or style. 

If there are no errors, please reply with the original sentence. Ensure that all the modifications you make are based on grammatical rules; do not embellish the sentence. Please format your final answer as follows.

\textless answer\textgreater

Your corrected sentence, or the original sentence

\textless /answer\textgreater

The sentence for you to modify is as follows:

[sentence]

\newpage

\begin{figure}[htbp!]
    \centering
    \includegraphics[width=0.9\linewidth]{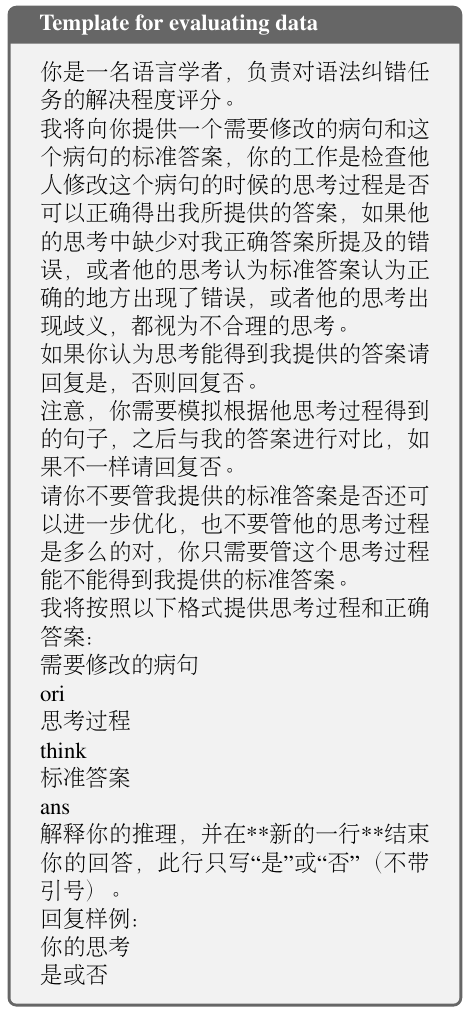}
    \caption{Template for evaluating data}
    \label{fig:pmt2}
\end{figure}

The English translation of fig\ref{fig:pmt2} is as follows:

You are a linguist responsible for evaluating the effectiveness of grammar correction.

I will provide you with a grammatically incorrect sentence that needs to be revised, along with the correct version of that sentence.
Your task is to determine whether another person's thought process for correcting the sentence is reasonable in terms of reaching the correct answer I provide.

If their thought process fails to identify any of the errors addressed in the standard answer,
or incorrectly identifies as wrong any part that the standard answer considers correct,
or their thought process contains ambiguity or confusion,
then it should be regarded as unreasonable.

If you believe that the thought process can produce the standard answer I provided, reply with Yes. Otherwise, reply with No.

Note: You must simulate the sentence that would result from the thought process, and then compare it to the provided standard answer.
If they are not the same, reply with No.

Do not consider whether my standard answer could be further improved,
nor how logically sound or elaborate the thought process is.
Your only concern is whether the thought process can lead to the standard answer I provided.

I will present the items in the following format:

Incorrect sentence

ori

Thought process

think

Standard answer

ans

Explain your reasoning, and on a new line, write only:

Yes or No (without quotes).

\newpage

\begin{figure}[htbp!]
    \centering
    \includegraphics[width=0.9\linewidth]{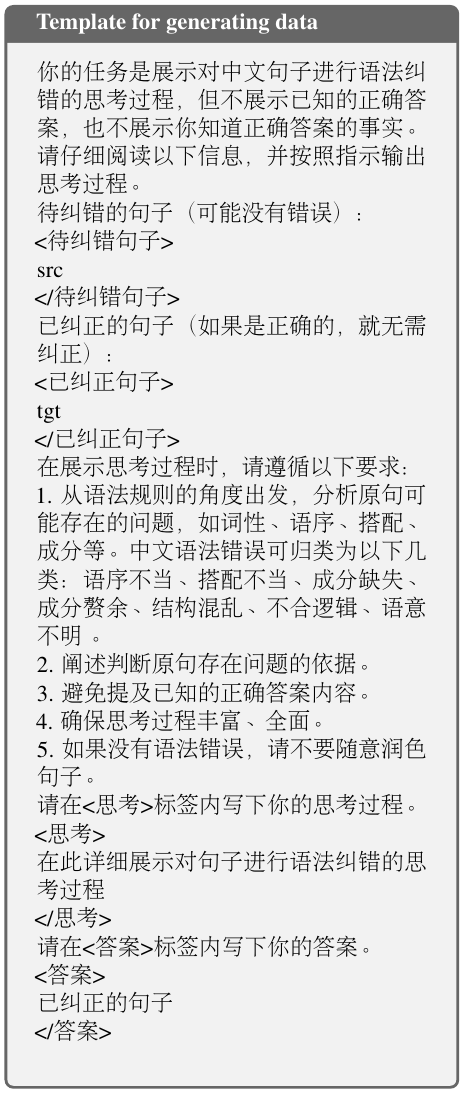}
    \caption{Template for generating data}
    \label{fig:pmt3}
\end{figure}

The English translation of fig\ref{fig:pmt3} is as follows:

Your task is to demonstrate the thought process for grammatically correcting a Chinese sentence, but without showing the known correct answer or revealing the fact that you know it. Please read the following information carefully and output the thought process as instructed.

Sentence to be corrected (may have no errors):

\texttt{<sentence\_to\_be\_corrected>}

src

\texttt{</sentence\_to\_be\_corrected>}

Corrected sentence (if it's already correct, no correction is needed):

\texttt{<corrected\_sentence>}

tgt

\texttt{</corrected\_sentence>}

When demonstrating the thought process, please follow these requirements:
1. From the perspective of grammatical rules, analyze potential problems in the original sentence, such as part of speech, word order, collocation, components, etc. Chinese grammatical errors can be categorized as follows: Improper word order, improper collocation, missing component, redundant component, confused structure, illogical, or ambiguous meaning.
2. Explain the basis for judging that the original sentence has a problem.
3. Avoid mentioning the content of the known correct answer.
4. Ensure the thought process is rich and comprehensive.
5. If there are no grammatical errors, do not stylistically polish the sentence.

Please write your thought process within the \texttt{<thought>} tags.
\texttt{}

Please write your answer within the \texttt{<answer>} tags.
\texttt{<answer>}
The corrected sentence
\texttt{</answer>}

\end{document}